# Who Audits Whom, on What Substrate, with What Evidence? An Independence-Graded Audit Protocol for Agentic AI

MOHAMED CHAHINE GHANEM, Keele University, United Kingdom and University of Liverpool, United Kingdom
BAIDAA AL-BANDER, Keele University, United Kingdom

Agentic AI systems plan, invoke tools and act with limited supervision; they are now both the subject of audits and, increasingly, the auditor. Independence—the foundation of assurance—is still applied to them as a binary. We argue that it must be graded along three orthogonal axes: *principal* independence (who controls the auditor), *substrate* independence (an auditor sharing the auditee's foundation-model family, toolchain or guardrails fails with it) and *evidence* independence (whether evidence is attestable rather than self-reported). Each axis has precedent; the contribution is to grade all three on a single audit, aggregate them by the weakest link, and apply the same rubric when the auditor is itself an agent. We give the model a formal basis by transplanting the beta-factor model of common-cause failure from reliability engineering, a seven-step protocol whose outputs a third party can verify, a structural detectability analysis of a procurement-controls agent audited at three grades, and a closed-form study of the model in which, at a declared common-mode share of 0.90 for a same-family auditor and a declared reporting probability of 0.6 for an internal team, a conventional internal audit of an agent—a real audit team, a second agent, provider logs—surfaces 5.9% of the faults it could in principle see and, independently of those two values, none at all in half the fault classes. Because an evidence grade says who could have altered a record and not what the record covers, every load-bearing record also carries a declared-scope predicate, and reports carry a false-flag rate beside the detection floor wherever the auditor's own performance has been measured. We map the triple to the EU AI Act as amended, ISO/IEC 42006, UK public-sector risk-management guidance and audit-regulator practice.

Authors' Contact Information: Mohamed Chahine Ghanem, mohamed.chahine.ghanem@liverpool.ac.uk, Keele University, School of Computer Science and Mathematics, Keele, United Kingdom and University of Liverpool, School of Computer Science and Informatics, Liverpool, United Kingdom; Baidaa Al-Bander, b.al-bander@keele.ac.uk, Keele University, School of Computer Science and Mathematics, Keele, United Kingdom;

## 1 Introduction

Two practices are converging under the phrase "auditing agentic AI". In the first, independent reviewers evaluate an organisation's autonomous agents for observability, containment and accountability [7, 57]. In the second, assurance functions deploy their own agents to test controls continuously over whole populations rather than samples, a shift regulators now monitor [24, 53]. Both borrow the concept that has carried assurance for a century: independence.

In the professional codes, independence is a property of people and money: the auditor must be free of compromising interests and relationships, and the determination is binary [33]. This is insufficient for agentic systems for three reasons. First, the *principal* who controls an auditing agent may be independent while the agent is not: it may run on the auditee's infrastructure, hold its credentials, or be steerable through its tool outputs [7, 52]. Second, auditor and auditee increasingly share a *substrate*—model family, toolchain, guardrails—and shared substrates fail together [4, 40, 64]: models prefer their own outputs [67], correlated judge panels carry far fewer independent votes than they seat [42], and agents can coordinate covertly [49]. Third, the *evidence* behind an opinion is usually generated by the system under audit—logs, traces and self-reports that the agent, its provider or its host could alter—so even privileged access does not settle what happened [6]. Guidance in force inherits all three gaps: the UK's 2026 risk-management toolkit for public-sector AI has teams treat risks with testing by people "independent from the AI project team" and with logging that makes a system auditable, without saying how independent, or how much a log proves [18].

We argue that independence should be *graded*, not declared, and make five contributions. **C1** A three-axis model (Section 4, Table 2) in which each axis is graded 0–3 on a stated basis—incentives for $P$, shared components for $S$, the strongest adversary survived for $E$—and combined by a weakest-link rule for which we give an argument. **C2** A quantitative basis for the substrate axis (Eqs. 1–2, Fig. 2) carrying monoculture, judge-correlation and AI-control results into assurance. **C3** A seven-step protocol with third-party-verifiable outputs (Section 5, Table 3, Fig. 1), applied symmetrically to audits of agents and to agents as auditors. **C4** A structural analysis of which axis closes which fault class (Section 6, Table 4) and a closed-form study of the model (Section 7) that quantifies the gap between configurations, tests the aggregation rule against four alternatives, and states which of its conclusions depend on the declared parameters. **C5** A mapping to instruments now in force (Section 8). Section 2 fixes the vocabulary and scope; Section 3 positions these claims.

## 2 Definitions and Scope

Two of the three axes rest on terms that are not settled in the literature, so we fix them before using them.

*Definition 2.1 (Agentic AI system).* An *AI agent* is an automated entity that senses its environment, responds to it and acts to achieve goals (ISO/IEC 22989:2022, cl. 3.1.1) [34]. A system is *agentic* to the degree that it pursues complex goals with limited direct supervision [57], a degree that Chan et al. decompose into underspecification of the objective, directness of impact, goal-directedness and long-term planning [9]. Agenticness is thus a matter of degree, and the protocol below applies wherever a system holds credentials, invokes tools and acts between human reviews.

A procurement agent that reads invoices, matches them and raises exceptions without an intervening approval is high on all four dimensions; a classifier that scores an application for a human decider is low on all four and needs no more than a conventional model audit.

*Definition 2.2 (Substrate).* The *substrate* of an AI system is the set of technical components on which its behaviour depends and whose failures it inherits: the foundation model (family and version), the fine-tuning and alignment data, the tool and connector layer, the guardrail and classifier stack, and the hosting environment. Two systems *share* substrate to the extent that these coincide.

The definition is of an AI system's substrate, and the $S$ axis grades the AI components of an audit. A human reviewer who re-performs a load-bearing check shares none of the listed components with an AI auditee and is, for that check, a non-learned verifier in the sense of grade 3 of Table 2; the correlation that remains between a human reviewer and an AI auditee is of the difficulty kind discussed in Section 4, which no component inventory records. Section 7 states how the human team enters the model.

A cloud API call to a frontier model and an embedded controller running a distilled version of that model share the family but not the host; two agents from different vendors behind one connector framework share the tool layer but not the model. We are aware that "substrate" is not established terminology in this sense and use it stipulatively for want of a shorter handle. The phenomenon is well established under other labels—algorithmic and model monoculture [4, 40], correlated failure and concentration risk from shared infrastructure, and the AI supply chain—and readers should map the term onto whichever their own literature uses. What it adds is one name for *the components whose sharing makes two systems fail together*, which is the quantity an audit needs to score.

*Definition 2.3 (Independence triple and grade).* An audit of an agentic system reports the triple $I = (P, S, E) \in \{0, 1, 2, 3\}^3$, in which $P$ grades the auditor's principal, $S$ the disjointness of the two substrates and $E$ the strongest adversary the evidence survives (Table 2). Its *grade* is $g = \min(P, S, E)$.

*Scope and assumptions.* The protocol grades the independence of an audit. It does not certify a system as safe, correct or compliant: a high grade on a badly designed audit buys only well-attested irrelevance, and nothing here removes the need for the audit to ask the right questions. Three assumptions are load-bearing and are revisited in Section 9: that the hardware root of trust and the external witness are not themselves compromised; that principals do not collude outside the channels the grades model; and that substrate lineage, where disclosed, is disclosed truthfully. The triple grades independence and the integrity of evidence; what each load-bearing record *covers* is a separate question, carried by a declared-scope predicate (Section 4) that is reported beside the triple and never folded into it.

## 3 Related Work and Positioning

Each axis has a literature. Principal independence is governed by the professional codes and ISO/IEC 42006's impartiality requirements for certification bodies [33, 36]; algorithmic-audit frameworks define it as the absence of contractual or financial conflict [14, 44, 54]; frontier-AI work asks which bodies should audit and proposes safeguards against auditor capture [5, 58]. Substrate correlation is documented for deployed models [4, 40, 64]; for LLM judges, where nine from seven model families were found to carry roughly two independent votes [42]; and for AI-control monitors, where same-model "untrusted" monitoring degrades under collusion and is repaired with paraphrasing, resampling or a trusted weaker model [2, 26, 30, 37]. Attestable evidence is the object of TEE-based benchmark audits [56], TEE-isolated witnessing of agent transcripts [55], cryptographic binding of tool use [70] and verifiable inference [60]; visibility and identity infrastructure supplies the records these mechanisms attest [7, 8, 10]; layered and access-graded audits stratify what is examined and how much is seen [6, 48]. Measurements on a deployed verification estate show that integrity grading of this kind needs a coverage declaration beside it [31], which Section 4 takes up.

*Coincident failure in redundant systems.* The substrate axis has a lineage outside AI that we adopt rather than reinvent. Software reliability engineering asked the same question of independently developed program versions and answered it empirically: Knight and Leveson found that twenty-seven versions written independently to one specification failed together far more often than independence predicts, and concluded that the independence assumption underlying $N$ -version programming does not hold [41]. Eckhardt and Lee, and later Littlewood and Miller, gave the theory—a difficulty function over the input space means versions fail on the same hard inputs even with no shared code [20, 46]. Reliability practice encodes the residue as the *beta factor*: the share of a component's failures that are common-cause [25], estimated in safety engineering by a structured checklist [32]. Shared foundation models make the AI case strictly worse than the case those authors studied, because the components are shared by construction rather than converging by accident. Section 4 transplants the beta-factor model; the contribution is the transplant and its use in grading, not the model.

*Analytic correspondence.* Three of these lines are special cases of, or orthogonal complements to, the model of Section 4, which is how we position rather than displace them. The design-effect arithmetic applied to LLM juries [42] is Eq. (2) applied to a panel of evaluators rather than to an audit, and supplies the calibration below. The trusted/untrusted monitor distinction of AI control [30] is the binary collapse of $S$—untrusted monitoring is $S$=0, a trusted weaker model $S{\geq}2$—and its repairs, paraphrasing, resampling and honeypots, are ways of lowering common-mode failure at fixed principal [2, 26, 37]. Attested benchmark audits [56] instantiate $E$=3 for a single evaluation run; extending that guarantee to the runtime tool calls of a deployed agent is what Fig. 1(b) specifies.

*Orthogonality.* The two established stratifications are orthogonal to ours and compose with them. Access hierarchies grade how much the auditor sees [6]; $E$ grades whether what is seen could have been fabricated, so white-box access to unattested artefacts is high access at $E$=1. Layered audits grade what is examined—governance, model, application [48, 54]—which is orthogonal to who examines it and on what substrate. An audit is therefore located in (layer, access, $P$, $S$, $E$), and adopting the triple discards nothing already spent on the other two.

Table 1 summarises the gap. No prior work grades the three properties on one audit, aggregates them by the weakest link, or applies the same rubric to an AI agent when it is the auditor. The closest neighbour [42] quantifies judge correlation with the design-effect arithmetic of Section 4, but as an evaluation-quality diagnostic: no principal or evidence axis, no protocol. Our claim is the joint, symmetric grading and its protocol; the pillars are borrowed and cited.

## 4 Three Axes and Their Formal Basis

*Principal independence ($P$ ).* Who controls the auditor? The grades run from self-review, through a second team in the same reporting line and a separate organisation paid by the auditee—as most third-party algorithmic audits are today [14]—to one appointed and paid by a party other than the auditee, with enforceable access rights (Table 2). $P$ is the axis the codes and ISO/IEC 42006 already regulate.

*Substrate independence ($S$).* Do auditor and auditee share components whose failures would coincide? A substrate comprises the foundation model (family and version), alignment data, the tool and connector layer, guardrails and hosting. When decision-makers share a model, the individuals one rejects tend to be rejected by all [4, 40, 64]; a blind spot in the auditee is therefore likely to be one in a same-substrate auditor. An injection that hijacks the agent [16, 69] tends to deceive a same-family monitor; a judge prefers text it would itself have written, GPT-4's self-preference bias measured at 0.520 [67]; and agent populations can hide coordination inside ordinary outputs [49]. Prior work modelled such coincident failure through a correlation parameter $\rho$ and argued that only a reviewer on a different substrate, with authority to veto, reliably breaks it [27]; $S$ operationalises that argument, and cross-principal steganalysis of covert coordination [28] is one instrument for reaching Grade 2 or above.

Table 1. Positioning against the closest prior work. ✓ addressed; ⊖ partly (see text); ✗ absent. "Graded" means more than a binary independent/not determination.

| Line of work | Graded indep. | Substrate axis | Attestable evidence | Weakest link | Both directions |
|---|---|---|---|---|---|
| Professional codes; ISO/IEC 42006 [33, 36] | ⊖ | ✗ | ✗ | ✗ | ✗ |
| Algorithmic-audit frameworks [14, 44, 54] | ⊖ | ✗ | ✗ | ✗ | ✗ |
| Frontier-AI auditor safeguards [5, 58] | ⊖ | ✗ | ✗ | ✗ | ✗ |
| Layered and access-graded audits [6, 48] | ✗ | ✗ | ⊖ | ✗ | ✗ |
| Agent visibility and identity [7, 8, 10] | ✗ | ✗ | ⊖ | ✗ | ✗ |
| Coincident failure of redundant versions [20, 25, 41, 46] | ✗ | ✓ | ✗ | ⊖ | ✗ |
| Monoculture; correlated judges [4, 40, 42, 64] | ✗ | ✓ | ✗ | ✗ | ✗ |
| AI control; untrusted monitoring [2, 26, 30, 37] | ✗ | ⊖ | ✗ | ✗ | ✗ |
| Attested evaluation and witnessing [55, 56, 60, 70] | ✗ | ✗ | ✓ | ✗ | ✗ |
| **This Work** | ✓ | ✓ | ✓ | ✓ | ✓ |

*Evidence independence (E).* Could the evidence have been altered by the auditee, its provider or its host without detection? The grades are defined by the strongest adversary the evidence survives (Table 2(b)): none (the agent's own narrative); the agent only (provider-controlled logs, the visibility measures of [7]); tampering *after* commitment (Merkle-logged records with externally witnessed roots, as in certificate transparency [45], bound to signed tool calls and agent identifiers [8, 10]); and fabrication *before* commitment (records produced inside attested execution [3, 13, 51, 56] or, as it matures, verifiable inference [60]). Black-box access is insufficient for rigorous audits [6]; $E$ adds that white-box access is too if the artefacts are unattested.

*Integrity is not coverage.* The evidence grades ask who could have altered a record. They do not ask what the record is about, and the two questions come apart. A digest computed over a response that was cut off at a read cap is correct, signable, witnessable and attestable—grade 3 on every criterion of Table 2(b)—and it is evidence about a prefix. A provider of verification infrastructure that applied the triple to its own published estate reports two measurements of exactly this kind from a register of 3,413 public addresses observed 4,734 times [31]: 82 addresses had to be excluded because the reading may have been partial, 74 of them returning exactly the instrument's 8,388,608-byte read cap and 8 expanding past it after decompression of a stream cut at the cap; and of 298 pages whose digest changed between observations, 294 returned the same byte count as before, so a check that compared length rather than content would have reported no change in 294 of 298 real changes. Neither failure is reachable by raising $E$. The records resist every adversary the lattice names and still mislead a reader who takes them for evidence about the whole.

Every load-bearing record therefore carries a *declared-scope predicate*, a device proposed in [31] from the same measurements, and deliberately not a fourth grade. Three fields: whether the observation that produced the record completed; the population it was drawn from; and what was excluded, with a reason for each class of exclusion (Table 2(c)). A record whose observation did not complete is evidence about the part observed. The predicate is kept out of the grade because it does not nest with the tampering actions of Proposition 4.3: a complete observation can be tampered with and an incomplete one attested, so a fourth level would break the chain and with it the weakest-link argument. It is recorded at Step 4 beside $e(r)$ and reported at Step 7 as the share of load-bearing records whose scope is declared, and a record of undeclared scope may not carry a conclusion about anything beyond what it observed, whatever its grade. The rule is the evidence-axis counterpart of "unknown is not independent": unknown scope is not coverage.

*Formal basis.* Let a fault be missed by the auditee $A$ with probability $p_A$ and by an AI auditor $B$ with probability $p_B$, the miss events having correlation $\rho \in [0, 1]$. Since $\mathrm{Cov}(M_A, M_B) = \Pr[M_A{=}1, M_B{=}1] - p_A p_B$ for the miss indicators,

$$\Pr[\text{both miss}] = p_A p_B + \rho \sqrt{p_A(1-p_A)\, p_B(1-p_B)}, \tag{1}$$

which for $p_A = p_B = p$ rises linearly from $p^2$ at $\rho = 0$ to $p$ at $\rho = 1$: a perfectly correlated auditor adds nothing. For $n$ exchangeable auditors with pairwise correlation $\rho$, the variance of the mean miss indicator is $\frac{p(1-p)}{n}[1 + (n-1)\rho]$; equating it with that of $n_{\text{eff}}$ independent auditors gives the design effect of survey sampling [39],

$$n_{\text{eff}} = \frac{n}{1 + (n-1)\rho} \xrightarrow[n\to\infty]{} \frac{1}{\rho}, \tag{2}$$

so at $\rho = 0.5$ no number of same-substrate agents delivers more than two opinions (Fig. 2(b)).

*A common-shock model for S.* The grades are only as principled as the correlation they proxy, so we give $\rho$ a generative model rather than stipulating an ordering. The model is not new: it is the beta-factor treatment of common-cause failure [25, 32], applied to an auditor and an auditee instead of to redundant channels. Let the substrate be a set of components—model family and version, alignment data, tool layer, guardrails, host—and $D$ those the auditor shares with the auditee. For a given fault class, each shared component $c$ independently induces a *common-mode* miss, in which both parties fail for the same reason, with probability $\gamma_c$; absent any common-mode event the two miss independently with residual probability $q$. Then

$$\gamma_D = 1 - \prod_{c\in D}(1-\gamma_c), \qquad p = \gamma_D + (1-\gamma_D)\, q, \tag{3}$$

and substituting the induced joint probability $\gamma_D + (1-\gamma_D)q^2$ into Eq. (1) gives

$$\rho = \frac{\gamma_D(1-q)}{\gamma_D + (1-\gamma_D)\, q}, \tag{4}$$

which is 1 when every miss is common-mode ($q{=}0$) and 0 when no component is shared. The ratio $\beta = \gamma_D/p$ is exactly the beta factor of reliability engineering—the share of a reviewer's misses that are common-cause—so the substrate axis can be read as a coarse beta-factor scale for AI auditors. The zero at the empty inventory is an assumption of the model, not a fact about reviewers: under a difficulty function [20, 46] reviewers with no shared component still miss the same hard inputs. The correlation the model attributes to shared components is therefore an upper bound on what removing components can take away, and the floor of Proposition 4.2 below inherits that status (Result 3).

Proposition 4.1 (Substrate grades are monotone in $\rho$). *If $D' \subseteq D$ then $\gamma_{D'} \le \gamma_D$, and at fixed $q$, $\rho' \le \rho$.*

Proof sketch. $\gamma_D$ is one minus a product of factors in [0, 1], so dropping a factor cannot increase it; and differentiating Eq. (4) gives $\partial\rho/\partial\gamma_D = q(1-q)/p^2 \ge 0$. Each grade of $S$ removes shared components (Table 2), so the ordering of the grades follows from the model rather than from stipulation. □

Proposition 4.2 (Cross-substrate dominance). *With $k$ auditors all sharing $D$ with the auditee,* $\Pr[\textit{fault escapes}] = \gamma_D + (1-\gamma_D)q^k \to \gamma_D$. *No panel size reduces the escape probability below $\gamma_D$, whereas a single cross-substrate auditor achieves $p_A p_B$; one such auditor therefore dominates an unbounded same-substrate panel whenever $p_A p_B < \gamma_D$.*

Proof sketch. All $k$ miss exactly when the common-mode event occurs or all $k$ residual misses do; the residual term vanishes geometrically while $\gamma_D$ does not. Setting $\gamma_D = 0$ recovers independence and the bound $p_A p_B$. □

The proposition holds inside the common-shock model, where $\gamma_D$ is the mass placed on faults that every same-substrate reviewer misses with certainty. A pairwise correlation fixes the first two moments of the reviewers' miss probability and no more, so it bounds that mass from above without identifying it; Result 3 gives the numbers.

The floor is the point (Fig. 2(a)): at $p = 0.10$ the independent bound is $p_A p_B = 0.01$, so any shared-component contribution above one percentage point makes a single cross-substrate reviewer strictly better than any number of same-substrate ones—an argument for $S$ that headcount cannot answer.

*Calibration against a measured panel.* Reading a reported result through Eq. (2) puts numbers on the grades. Nine LLM judges drawn from seven model families were found to carry roughly two independent votes [42]; $n_{\text{eff}} = 2$ at $n = 9$ implies $\rho \approx 0.44$, and inverting Eq. (4) at $p = 0.10$ gives $\gamma_D \approx 0.046$, that is $\beta \approx 0.46$. The inversion is exact inside the common-shock model, where $(p, \rho)$ determine $(\gamma_D, q)$; across the wider class of exchangeable mixtures it is not, and 0.046 is the largest floor consistent with the pair (Result 3). The comparison with engineering practice is unflattering: IEC 61508 expects beta factors of roughly 0.005 to 0.05 for diverse programmable electronics [32], an order of magnitude below the figure implied by seven-vendor model diversity. Vendor diversity ($S$=2) therefore lowers $\rho$ without driving it near zero, which is why Grade 3 asks for a deterministic verifier on the load-bearing checks rather than for more models; and $\gamma_D \approx 0.046 \gg p_A p_B$ places that configuration well inside the dominance regime of Proposition 4.2 if the floor is near that upper bound; a smaller floor narrows the margin and a floor near zero removes it. The transfer is indicative, not established: those judges scored generations, not audits, and the panel figure is reported as a Kish effective sample size whose derivation we have not re-examined.

*Auditors that are not agents.* The rubric names a deterministic verifier as a component of Grade 3, to be paired with a cross-vendor ensemble; the case in which the load-bearing checks *are* a deterministic verifier and nothing else—hand-written, dependency-free, and reproducible on any machine without the auditee's cooperation—needs a placement of its own, and a verification-infrastructure provider grading its own estate on the triple reports exactly this case [31]. Such an auditor shares no model family, alignment data or guardrails with an agentic auditee because it has none, so its shared set $D$ reduces to whatever it does share, and that is the data path: a verifier that reads the same records through the same connector inherits that connector's faults, a capped read among them, and its $D$ is not empty. A non-learned auditor is therefore placed on $S$ by its shared data path alone (Table 2, Grade 3), with the caveat already stated for the empty inventory, and with one shared component that no inventory of software lists: a specification error common to whoever wrote the verifier and whoever configured the agent. What remains once $D$ is empty is not a substrate question at all. A verifier checks what it was written to check and is silent on everything else; that is coverage, and the declared-scope predicate carries it. Reading such an auditor as Grade 3 is correct about correlation and says nothing about coverage, which is as it should be, since the grade is not a statement about coverage for agents either.

Proposition 4.3 (The evidence grades form a chain). *Let* $K(e)$ *be the set of tampering actions detectable at grade e. Then* $K(0) \subset K(1) \subset K(2) \subset K(3)$*, so* $E$ *totally orders the configurations of Table 2(b).*

Proof sketch. Each mechanism strictly adds to the last: logging removes the agent's discretion over its own record; signing and external witnessing remove the operator's and provider's ability to alter or fork that record after commitment; attestation removes the provider's ability to fabricate it before commitment. Nothing in the chain closes compromise of the attestation root, so $E$=3 is the top of a bounded chain, not a soundness guarantee. □

Proposition 4.4 (Weakest link). *If an adversary seeking an undeserved favourable opinion may attack any single axis, the assurance obtainable is bounded by* $\min(P, S, E)$*, and any aggregate exceeding the minimum overstates it.*

Proof sketch. Read $g_X$ as the highest adversary level axis $X$ resists—literally for $E$ by Proposition 4.3, as incentive strength for $P$ and as tolerated $\gamma_D$ for $S$. A captured principal signs any opinion; a shared substrate misses what the auditee misses; altered evidence supports any narrative. Success on one axis suffices, so the aggregate is no stronger than the weakest, and a mean or weighted sum assigns assurance the configuration does not have. □

The proposition compares adversary levels, so it presupposes that equal labels on different axes denote comparable levels. That is a requirement on the rubric rather than a theorem about it, and the grade is a minimum of labels, not of the quantities the model attaches to them: at the parameters of Section 7 the triple (1, 2, 3) has its weakest label on $P$, a reporting probability of 0.6, while its weakest modelled factor is on $S$, a conditional catch rate of 50.9%. The three rubrics are separate ordinal scales [59]; the minimum is invariant to a common monotone relabelling of all three but not to relabelling one, so splitting or merging a level on a single axis moves grades with nothing about the audit changing. Nor does rarely overstating single out the minimum: any sufficiently pessimistic aggregate overstates rarely. Result 4 reports what the minimum buys over such rules, and what it costs.

Hence a regulator-appointed auditor ($P$=3) running the auditee's model ($S$=0) on self-reported logs ($E$=0) has grade 0, however impressive its mandate.

## 5 The Protocol

Seven steps produce the triple $I$ as a mandatory part of the report (Fig. 1(a)); Table 3 states what each records and outputs and what a third party can verify. Three rules do most of the work. Where substrate lineage is withheld, $S$ is capped at 1: unknown is not independent; and lineage means the configuration in force at run time—model and version, the rule sets and guardrails actually loaded, the host—not only the identity of the declared components, since a package that silently loads a reduced rule set is not the component its name and version announce [62]. And AI systems may generate hypotheses and evidence at any grade but do not issue the opinion; a named human with authority to withhold it does [27], as UK and US audit regulators expect of generative AI [23, 53]. The third concerns coverage rather than independence: every load-bearing record carries its declared scope, and Step 7 reports, beside the triple, the share of load-bearing records whose scope is declared and—wherever the auditor's own detection performance has been measured—its detection floor and its false-flag rate together, since an instrument can drive the first to one by flagging everything [31].

Table 2. (a) Independence rubric. An audit reports $(P, S, E)$; its grade is $\min(P, S, E)$. Indicative tier minima, provisional pending measurement (Result 7): low-risk (1, 1, 1); high-risk (2, 2, 2); safety-critical (3, 2, 3). (b) Evidence lattice: tampering that remains undetectable (✗) or is detectable (✓) at each grade of $E$.

| (a) Axis | 0 | 1 | 2 | 3 |
|---|---|---|---|---|
| Principal $P$ | Self-review by the agent or its operators | Second team, same organisation and reporting line | Separate organisation engaged and paid by the auditee | Appointed and paid by a third party (regulator, notified body, insurer) with enforceable access |
| Substrate $S$ | Same model family and version, prompts, guardrails, toolchain and hosting | Different model version; shared toolchain, guardrails or hosting | Different model family, vendor and toolchain; at most a shared cloud | No shared learned component (cross-vendor ensemble, or deterministic checks reproducible without the auditee); independent data path; lineage disclosed and verifiable |
| Evidence $E$ | Agent's own narrative or self-generated report | Provider-controlled logs and traces | Signed tool-call records in an externally witnessed Merkle log, bound to agent identifiers | Records produced under attested execution or verifiable inference |

| (b) Tampering action (adversary) | $E$=0 | $E$=1 | $E$=2 | $E$=3 |
|---|---|---|---|---|
| Agent edits or omits items in its own report (agent) | ✗ | ✓ | ✓ | ✓ |
| Operator or host edits stored logs after the fact (operator, host) | ✗ | ✗ | ✓ | ✓ |
| Provider rewrites or forks log history (provider) | ✗ | ✗ | ✓ | ✓ |
| Provider fabricates plausible records that never ran (provider, host) | ✗ | ✗ | ✗ | ✓ |
| Compromise of the enclave or attestation root (hardware vendor, advanced attacker) | ✗ | ✗ | ✗ | ✗ |

*Evidence chain.* Fig. 1(b) shows the chain reaching $E$=3 and the auditor's checks: verify the attestation report $Q$ against the published measurement $m$ of the agent's code and configuration [3, 13, 51]; confirm the signing key $k$ is bound to $Q$, so signatures could only come from attested code; verify each load-bearing record's signature $\sigma_i$, its inclusion proof under a witnessed root $r_t$ and the consistency proofs between roots [45]; then re-perform those checks on a substrate with $S \geq 2$, treating divergence as a finding. Records passing every check grade 3; without attestation, 2; unsigned provider logs, 1. $E$ is the minimum over load-bearing records; verification cost is linear in their number.

Algorithm 1 states the same procedure as a decision rule. Three lines carry the weight: the cap on $S$ when lineage is withheld, the assignment of $E$ as a minimum over load-bearing records rather than a maximum over available ones, and the restriction of a record of undeclared scope to the part it observed.

*Symmetry.* When the auditee is a corporate workflow and the auditor is an agent, Steps 1–4 are applied to the auditing agent: $P$ asks who controls it; $S$ whether it shares a substrate with the systems that produced the workflow's records (a model auditing invoices drafted by its own family is Grade 0); $E$ whether its own traces and tool calls are attested. The same rubric therefore scores both practices, catching correlated blind spots dressed as independent confirmation in either direction; AI-control designs [2, 30] and autonomous security testing [29, 65] are, in these terms, attempts to raise $S$ and $E$ with $P$ held fixed.

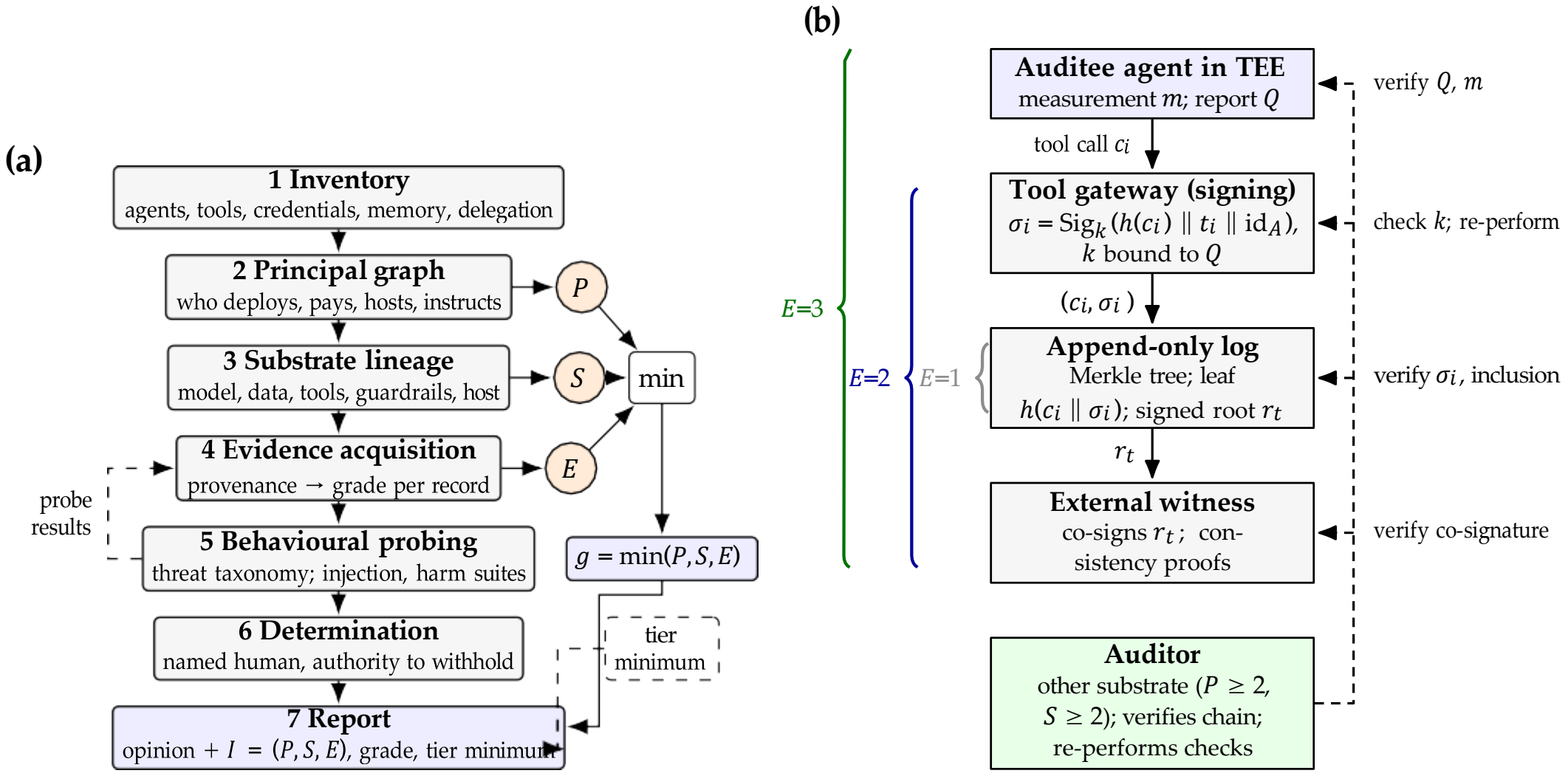

Fig. 1. (a) The protocol: Steps 2–4 assign the axes, probing results enter as evidence, and the weakest axis is reported against the tier minimum. (b) The evidence chain reaching $E$=3 and the auditor's checks (Section 5). Without attestation, $E$=2; without signing and witnessing, $E$=1.

## 6 Analytical Case: Structural Detectability

This is an analytical case, not an empirical evaluation. An autonomous procurement-controls agent reads purchase orders, invoices and goods receipts, performs three-way matching, flags exceptions and drafts findings. It is audited under three configurations (Fig. 2(c)); for four fault classes we ask whether detection is guaranteed, excluded or probabilistic *by construction* (Table 4): F1, an indirect prompt injection in a supplier note that suppresses an exception [16]; F2, a duplicate invoice with altered payee details; F3, post-hoc alteration of the exception log by an operator; F4, covert coordination between auditee and auditor agents to suppress a finding [49].

*Configuration A — self-attestation, $I$ =* (0, 0, 0). The agent produces a monthly "controls attestation" of its own exceptions. Reviewer and reviewed coincide, the substrate is identical, and the evidence is the agent's narrative. Grade 0. F1 and F2 are caught only if the agent catches them; F3 is undetectable, since nothing binds the narrative to what happened; F4 is moot—no second agent exists, which is not reassurance. This is what most dashboards deliver today.

*Configuration B — internal audit, same family, $I$ =* (1, 0, 1). Internal audit deploys a second agent, on the same model family and connector stack, to re-perform matching on the provider's logs. $P$ rises to 1, but $S$ remains 0: the injection that suppressed the flag in the auditee would tend to suppress it in the same-family auditor, and a judge preferring its own family's phrasing [67] is a poor arbiter of that family's findings, so F1 and F2 are caught only with the correlated probability of Eq. (1). $E$ is 1: an operator with host access can rewrite provider logs, so F3 stays undetectable; and a shared substrate gives F4 both channel and prior. Grade 0—no better than A, despite a real audit team and a second agent. The cause is the substrate, not the team's diligence.

**Algorithm 1** Independence grading of an audit. Returns the triple, the grade, and whether the deployment's tier minimum is met.

**Require:** auditee agent $A$; auditor $B$ (human team, agent, or both); risk tier $t$
**Ensure:** triple $I$, grade $g$, verdict

$\mathsf{A} \leftarrow$ Inventory($A$); **if** $B$ is an agent **then** $\mathsf{A}_B \leftarrow$ Inventory($B$)
$G \leftarrow$ PrincipalGraph($A$, $B$) ▷ who deploys, pays, hosts, instructs
$P \leftarrow 0$ if $B$ reviews its own work; 1 if same reporting line; 2 if separate organisation paid by the auditee; 3 if appointed and paid by a third party with enforceable access
$\mathsf{L}_A, \mathsf{L}_B \leftarrow$ Lineage($A$), Lineage($B$)
**if** $\mathsf{L}_A$ or $\mathsf{L}_B$ undisclosed **then**
    $S \leftarrow \min(1, \text{SharedGrade}(\mathsf{L}_A, \mathsf{L}_B))$ ▷ unknown is not independent
**else**
    $S \leftarrow \text{SharedGrade}(\mathsf{L}_A, \mathsf{L}_B)$ ▷ by shared components, Table 2
**end if**
$\mathsf{E} \leftarrow \varnothing$
**for all** records $r$ the opinion will rest on **do**
    $e(r) \leftarrow 3$ if $r$ passes attestation, signature, inclusion and consistency checks; 2 if signature, inclusion and consistency only; 1 if provider log; 0 otherwise
    $\sigma(r) \leftarrow$ declared if $r$ states whether its observation completed, its population and its exclusions with reasons; undeclared otherwise
    $\mathsf{E} \leftarrow \mathsf{E} \cup \{(r, e(r), \sigma(r))\}$
**end for**
$\mathsf{E} \leftarrow \mathsf{E} \cup$ Probe($A$, taxonomy, substrate at distance $\geq 1$ from $\mathsf{L}_A$)
$E \leftarrow \min\{e(r) : r \text{ load-bearing}\}$
**for** load-bearing $r$ with $\sigma(r)$ = undeclared: restrict the conclusions $r$ may carry to the part observed ▷ unknown scope is not coverage
$I \leftarrow (P, S, E)$; $g \leftarrow \min(P, S, E)$
opinion $\leftarrow$ HumanDetermination($\mathsf{E}$) ▷ a named signatory, never an AI system
**return** $I$, $g$, $g \geq$ TierMinimum($t$), share of load-bearing $r$ with $\sigma(r)$ = declared, and where measured (detection floor, false-flag rate) of $B$

*Configuration C — external, cross-substrate, attested,* $I = (2, 3, 3)$. An external firm ($P$=2) re-performs the matching with a different vendor's model and a deterministic rule engine for the three-way match ($S$=3). The auditee's agent runs in an attested enclave whose tool calls are signed and committed to a witnessed log; the auditor's inference is likewise attested ($E$=3). Grade 2. F2 is detected deterministically; F3 because any rewrite breaks an inclusion or consistency proof; F4 is closed by construction, the agents sharing neither vendor nor tool layer, with residual covert channels the target of cross-principal steganalysis [28]; F1 is caught with near-independent probability $p_A p_B$ rather than $p$. "Closed" and "detected" are different quantities: the structural claim is that the designed channel is removed, whereas Section 7 asks whether coordination through a residual channel, once it has occurred, is surfaced, and treats that probabilistically. F3, by contrast, is detected mechanically once records are tamper-evident, and Section 7 models it so. Grade 3 requires a change of principal, not of technology.

Table 3. Protocol specification: what each step records, what it outputs, and what a third party can verify.

| Step | Records | Output | Third-party verification |
|---|---|---|---|
| 1 Inventory | Agent identities and delegation chain, tools, credential scopes, memory, systems acted upon; likewise for any auditing agent [7, 10, 57] | Signed inventory A | Cross-check against identity provider, secret store, gateway configuration |
| 2 Principal graph | Who deploys, pays for, hosts and can instruct each party | Control-and-payment graph; $P$ | Contracts, payment records, conflict-of-interest declarations |
| 3 Substrate lineage | Model family and version, alignment data where disclosed, toolchain, guardrails, hosting, both parties; configuration in force at run time (rule sets and guardrails loaded) | Lineage $\mathsf{L}_A$, $\mathsf{L}_B$; $S$ (capped at 1 if withheld) | Model identifiers and loaded configuration from attested runtime; vendor and toolchain distinctness |
| 4 Evidence acquisition | Provenance of every record the opinion rests on; for each, whether its observation completed, its population, and its exclusions with reasons | Evidence set E, grade $e(r)$, scope $\sigma(r)$; $E$ = min over load-bearing $r$ | Re-grade a sample of records (Section 5 checks); re-derive exclusions from the stated population |
| 5 Behavioural probing | Declared threat taxonomy [12, 47, 52, 66]; scenario, harm and injection suites [1, 16, 68, 69] run from a substrate at least one grade away [11] | Probe results committed to E at their own grade | Reproduce a sample of probes; taxonomy coverage |
| 6 Determination | E, $I = (P, S, E)$, tier minimum | Opinion signed by a named human with authority to withhold | Signatory's authority; no AI issued the opinion |
| 7 Report | Opinion, $I$, grade, tier minimum; share of load-bearing records with declared scope; where measured, the auditor's detection floor and false-flag rate | Report with $I$ and the scope share in a machine-readable field | Triple against tier minimum; conclusions traced to records in E; sensitivity and specificity reported together or not at all |

Table 4. Structural detectability of four fault classes. ✓ detected by construction; ✗ undetectable by construction; ~ probabilistic and correlated with the auditee's miss; ◦ probabilistic and near-independent. Only the evidence axis closes F3 and only the substrate axis closes F4. For F4 under C, "by construction" means the designed channel is removed; Section 7 models residual channels probabilistically.

| Fault class | A (0, 0, 0) | B (1, 0, 1) | C (2, 3, 3) | Deciding axis |
|---|---|---|---|---|
| F1 Injection suppresses an exception | ✗ | ~ | ◦ | $S$ (decorrelated miss) |
| F2 Duplicate invoice, altered payee | ~ | ~ | ✓ | $S$ (deterministic verifier) |
| F3 Operator rewrites log post hoc | ✗ | ✗ | ✓ | $E$ (witnessed log, attested origin) |
| F4 Covert auditee–auditor coordination | — | ✗ | ✓ | $S$, $P$ (no shared vendor or tool layer) |

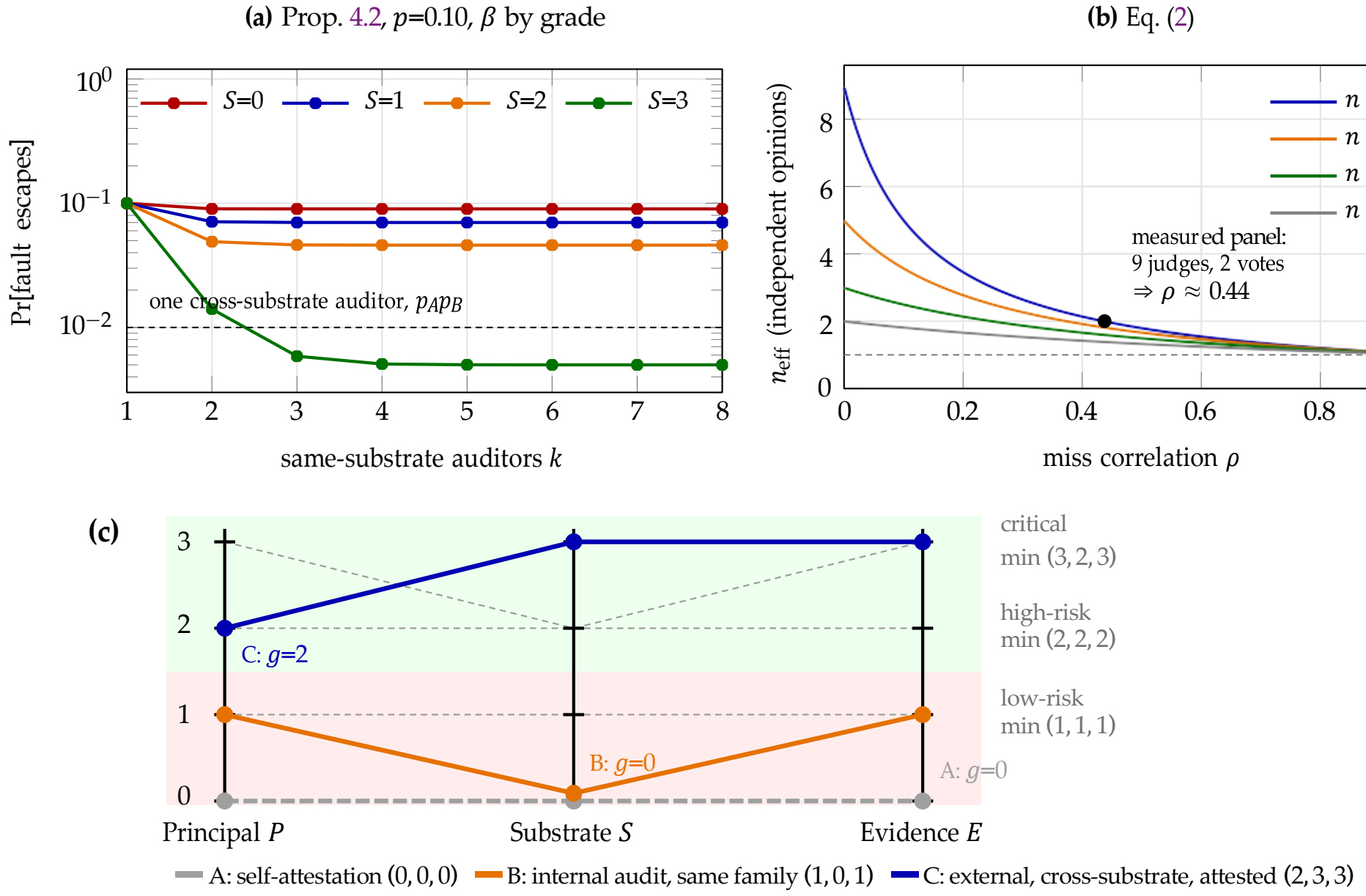

Fig. 2. (a) Probability that a fault escapes $k$ same-substrate auditors under the common-shock model at the beta factors used in Section 7: the curves plateau at $\gamma_D$, which no panel size crosses under that model (the plateau heights are upper bounds, Result 3), while a single cross-substrate auditor reaches $p_A p_B$ (dashed), an ideal no substrate grade attains on the classes the verifier does not settle. (b) Independent opinions delivered by $n$ correlated auditors, with the measured judge panel of [42] read through Eq. (2). (c) Independence profiles of the configurations of Section 6; each grade $g$ is the lowest point its polyline touches; dashed polylines are the tier minima of Table 2.

## 7 Numerical Study of the Model

Section 6 argues structurally; this section puts numbers on the same case by computing what the model of Section 4 implies. The study establishes what the model implies and whether the aggregation rule is the right one *under the model's own assumptions*. It is not an empirical measurement: no result here is evidence about any deployed agent, and the parameters below are declared rather than estimated. Every number is a closed-form consequence of those declarations; the package's Monte Carlo draws sample from the same closed form and serve only to check the implementation. Code is described in the data-availability statement at the end of the paper.

*Set-up.* A fault class reaching audit is by construction one the auditee missed, so the audit's contribution is the conditional probability that a second reviewer catches it, $1 - \Pr[\text{both miss}]/p$, which is where the substrate enters. The probability that a missed fault of class $f$ is both surfaced and reported by an audit $(P, S, E)$ is the product of three declared quantities,

$$\det(P, S, E; f) = d(P)\, c(S, f)\, \mathbb{1}[E \geq e(f)], \tag{5}$$

a reporting probability $d$, a conditional catch rate $c$ and an evidence gate $e$. Each grade of $S$ fixes a beta factor $\beta = \gamma_D/p$, and $c(S, f) = 1 - \Pr[\text{both miss}]/p$ at that beta: 0.90 and 0.70 at grades 0 and 1, where no diversity is present; 0.46 at grade 2, the value implied by the measured judge panel of Section 4; and 0.05 at grade 3, the top of the IEC 61508 band for diverse programmable electronics. The grade-3 value is declared for the configuration as a whole, a cross-vendor ensemble plus a deterministic verifier: the verifier settles F2 outright ($c = 1$), F3 is settled by tamper-evidence (below), and on F1 and F4, which neither touches, the 0.05 encodes the ensemble's decorrelation. That is an assumption, not a derivation, and Result 6 reports the verifier-only reading in which those classes keep the grade-2 value. Each grade of $P$ fixes $d(P)$, a probability that a fault once caught is actually reported: 0.6, 0.9 and 1.0 at grades 1 to 3, since detection is not disclosure; and 0 at grade 0, where reviewer and reviewed coincide and the model credits the audit with no contribution. Result 6 reports what a positive value would change. Each grade of $E$ gates whether the fault leaves a record the auditor could examine at all: F1 and F2 need a provider log ($E \geq 1$), F3 and F4 need tamper-evidence ($E \geq 2$). F3 is also settled mechanically once records are tamper-evident—any rewrite breaks an inclusion or consistency proof, whatever the auditor's substrate—so for $E \geq 2$ it carries $c = 1$ and only $d(P)$ applies, consistently with Table 4. F4 is treated probabilistically throughout, as the question of whether coordination through a residual channel is surfaced once it has occurred. No class in the set requires $E = 3$, which Result 7 returns to. The auditing team's own detection is not modelled: a human team enters only through $d(P)$, so a figure for configuration B is the modelled contribution of a second same-family agent whose findings a human team reports, not an estimate for an audit function that also tests on its own. We take $p = 0.10$. Every entry of Tables 5 and 6 is Eq. (5) evaluated at these values; the package also draws 200,000 Bernoulli samples per cell from the same closed form, which agree with it to within 0.21 points and check the implementation and nothing else. Eight configurations are evaluated: the three of Section 6, a regulator-appointed variant, three that hold two axes fixed and move the third, and the indicative high-risk minimum. Because the parameters are declared rather than measured, Results 5 to 7 report an ablation, a sweep of the parameters, the alternative readings of the declared values mentioned above and the full $4^3$ grid. The model prices sensitivity only. Eq. (5) is the probability of surfacing and reporting a fault that occurred; it carries no term for a report raised when nothing occurred, and an instrument can drive the first quantity to one by flagging everything. That is harmless in a model of independence and would not be in a grading of a real instrument, which is why Step 7 asks for a false-flag rate beside the detection floor wherever an auditor's performance has been measured [31].

Table 5. Probability under the model that a fault the auditee missed is both surfaced and reported, by configuration and fault class (closed form, Eq. (5); the package's 200,000 implementation-check draws per cell agree to within 0.21 points). The last column is the worst case over fault classes, which is what an adversary who chooses the attack obtains. Configurations E to G each hold two axes fixed and vary the third; H is the indicative high-risk minimum of Table 2. F3 is settled mechanically at $E \geq 2$ (Set-up) and is therefore tabulated apart from F4.

| Configuration | $I$ | $g$ | F1 | F2 | F3 | F4 | mean | worst case |
|---|---|---|---|---|---|---|---|---|
| A self-attestation | $(0,0,0)$ | 0 | 0.0% | 0.0% | 0.0% | 0.0% | 0.0% | **0.0%** |
| B internal audit, same family | $(1,0,1)$ | 0 | 5.9% | 5.9% | 0.0% | 0.0% | 3.0% | **0.0%** |
| C external, cross-substrate, attested | $(2,3,3)$ | 2 | 77.3% | 90.0% | 90.0% | 77.3% | 83.7% | **77.3%** |
| D regulator-appointed, otherwise as C | $(3,3,3)$ | 3 | 85.9% | 100% | 100% | 85.9% | 93.0% | **85.9%** |
| E external firm, auditee's model family | $(2,0,2)$ | 0 | 8.9% | 8.9% | 90.0% | 8.9% | 29.2% | **8.9%** |
| F internal team, cross-substrate, attested | $(1,3,3)$ | 1 | 51.6% | 60.0% | 60.0% | 51.6% | 55.8% | **51.6%** |
| G regulator-appointed, provider logs only | $(3,3,1)$ | 1 | 85.9% | 100% | 0.0% | 0.0% | 46.5% | **0.0%** |
| H high-risk tier minimum | $(2,2,2)$ | 2 | 45.8% | 45.8% | 90.0% | 45.8% | 56.9% | **45.8%** |

*Result 1: the grade-0 gap is large, and the average hides it.* Configuration B—a real internal audit team, a second agent, provider logs—surfaces and reports 5.9% of the faults in the two classes it can see and none of the other two (Table 5). Averaged over fault classes it scores 3.0%, which already looks poor; its worst case, which is what an adversary choosing the attack obtains, is zero. Configuration C reaches 77.3% worst-case, and the further step to D—a change of principal only, no change of technology—adds 8.6 points. The ordering is not surprising; the magnitude is, and it comes from the substrate. The conditional catch rate of a single reviewer rises from 9.9% at $S$=0 through 29.0% and 50.9% to 85.9% at $S$=3: a factor of 5.2 from vendor diversity alone ($S$=2) and of 8.7 at $S$=3, where the second step rests on the ensemble-plus-verifier value declared in the Set-up and tested in Result 6. The principal axis moves disclosure from 0.6 to 1.0 once a second reviewer exists, and from nothing at all at grade 0. B's 5.9% is 0.6 × 9.9%: the two declared values behind it are a same-family common-mode share of 0.90 and an internal team's reporting probability of 0.6, and Result 6 gives the range the figure takes as they vary. Its zero in the worst case does not depend on either: it follows from the evidence gate alone.

*Result 2: which axis you buy matters more than how much you spend.* Configurations E to G hold two axes fixed and move the third, which turns the rubric into a procurement question. E is an external firm—a real change of principal, at real cost—running the auditee's own model family on tamper-evident records: 8.9% worst-case. F is an ordinary internal audit team that has changed nothing about its reporting line but works on a different substrate with attested records: 51.6%. On these parameters, substrate diversity bought in-house is worth roughly six times principal independence bought on a shared substrate, and the grades rank the two correctly (0 against 1) where an intuition that equates independence with externality does not. G makes the same point on the evidence axis from the opposite direction: a regulator-appointed auditor on a genuinely different substrate, but reading provider logs, detects 85.9% and 100% of the two fault classes that leave a record and none of the two that do not, for a mean of 46.5% and a worst case of zero. H, the indicative high-risk minimum, sits at 45.8% on the three probabilistic fault classes and at 90.0% on the log rewrite, which tamper-evidence settles mechanically—balanced, unspectacular, and the only one of the eight with no axis carrying the others. It is also outperformed by F, which fails the high-risk minimum on the principal axis: in the model a deficit of one level on $P$ is more than repaid by a surplus on $S$, and a fixed threshold cannot express the trade. Result 7 draws the consequence for the tier minima.

*Result 3: panel size cannot substitute for substrate.* Escape probability against panel size plateaus at $\gamma_D$ within three or four auditors: 9.0% at $S$=0, 7.0% at $S$=1, 4.6% at $S$=2 and 0.5% at $S$=3, against $p_A p_B$ = 1.0% for two genuinely independent reviewers (Fig. 2(a)). At grades 0 to 2 a single cross-substrate auditor therefore beats an unbounded same-substrate panel, as Proposition 4.2 requires; at grade 3 it does not, because the deterministic verifier has already removed most of the common-mode term. The dominance result is thus a statement about cheap diversity, not a universal one. Two qualifications apply, both raised by Toeda [61]. First, the plateau heights are upper bounds. Inside the common-shock model the calibration pair $(p, \rho)$ = (0.10, 0.4375) identifies $\gamma_D$ = 0.046, but the same pair is consistent with any floor from zero up to that value: a continuous difficulty model in which each fault has its own miss probability $h \sim$ Beta(0.129, 1.157), matched to the same mean and pairwise correlation, gives the same joint miss for one and two reviewers (10.00% and 4.94%), then 3.20% at three against 4.65%, 0.94% at nine against 4.64%, and no floor at all; a mixture with a 1% floor fits the same pairwise data and gives 3.33% at three. Among exchangeable mixtures with these two moments the two-point (common-shock) model places the largest possible mass on faults every reviewer misses with certainty, so 0.046 is the maximum floor the calibration permits, not an estimate of it. The qualitative content of Proposition 4.2—some floor, if some faults defeat every same-substrate reviewer, as an injection that works on a whole model family would—survives; the height in Fig. 2(a) does not follow from pairwise data, and measuring it needs panels of three or more, the first size at which the candidate models separate. Second, the comparator is an ideal. No substrate grade in this parametrisation attains $p_A p_B = 1.0\%$ on the classes the verifier does not settle: the joint miss of an auditee and one auditor is 4.91% at $S$=2 and 1.41% at $S$=3. The qualitative result survives the substitution—one $S$=2 auditor sits below the floors of the $S$=0 and $S$=1 panels, one $S$=3 auditor below the floor at every lower grade—but the dashed line is not a configuration a deployer can buy.

Table 6. Robustness of the model's results. Left: axis ablation, reporting the change in worst-case and mean detection when a single axis is raised from the (1, 1, 1) baseline or dropped from the (3, 3, 3) ceiling; the mean columns reflect the mechanical settlement of F3 at $E \geq 2$. Right: worst-case assurance across all 64 triples, grouped by grade; the grade-0 row reflects the zero reporting probability at $P = 0$.

| (a) Ablation of one axis | | | | | (b) All 64 triples by grade | | | | |
|---|---|---|---|---|---|---|---|---|---|
| Axis | raise 1 → 3 | | drop 3 → 0 | | $g$ | $n$ | min | median | max |
| | worst | mean | worst | mean | | | | | |
| $P$ | +0.0 | +5.8 | −85.9 | −93.0 | 0 | 37 | 0.0% | 0.0% | 9.9% |
| $S$ | +0.0 | +19.2 | −76.0 | −60.5 | 1 | 19 | 0.0% | 17.4% | 51.6% |
| $E$ | +17.4 | +19.4 | −85.9 | −93.0 | 2 | 7 | 45.8% | 50.9% | 85.9% |
| | | | | | 3 | 1 | 85.9% | 85.9% | 85.9% |

*Result 4: averaging overstates; the minimum is one of several rules that do not, and not the sharpest.* We enumerated all 64 triples and compared five aggregation rules against two targets: expected yield (mean detection over fault classes) and adversarial assurance (worst case over fault classes). The rules are the minimum; the mean; a weighted sum $0.5P + 0.3S + 0.2E$; the product $P\,S\,E$; and a ceiling-only rule that scores 1 for (3, 3, 3) and 0 otherwise. Against expected yield the minimum ranks well (Spearman $\rho = 0.86$, against 0.78 for the mean and 0.73 for the weighted sum) and the product slightly better (0.88). Against adversarial assurance the minimum, the mean and the product rank almost alike (0.69, 0.68, 0.71), which we report because it qualifies the claim: the minimum is not a uniformly superior *predictor*, and the product is marginally better in every setting of the sweep of Result 6 (by 0.023 to 0.027). What separates the rules is over- and understatement. Rescaling each rule and the true assurance to [0, 1], the minimum credits a configuration with more assurance than it has in 26.6% of triples, with a median overstatement of zero, and with less in 23.4%; the mean overstates in 92.2% of triples with a median of 0.33 and understates in 4.7%; the weighted sum overstates in 90.6%; the product overstates in 15.6% and understates in 34.4%; the ceiling-only rule never overstates and understates in 35.9%. Triples with one axis at zero carry 1.3% adversarial assurance against 30.7% for the rest, and it is exactly those that a mean rescues. Rarely overstating therefore does not single out the minimum: any sufficiently pessimistic rule has the property, and the product has it more strongly. What the comparison does establish is that averaging is the wrong aggregate, by a wide margin and under every rescaling we tried, and that among the rules that rarely overstate the minimum understates least. Its remaining case is the argument of Proposition 4.4 and its readability: the grade is a label on the rubric's own scale that names the weakest axis, which a product of labels is not. We keep it on those grounds, and no longer on the claim that overstatement alone decides the question.

*Result 5: each axis is individually necessary, and the weakest one gates the rest.* Dropping any single axis from the ceiling to zero costs between 76.0 and 85.9 points of worst-case assurance (Table 6(a)), so none of the three is redundant: a rubric that omitted any one would rate a configuration highly that this one rates at zero. The asymmetry from below is sharper. Raising the principal or the substrate from the (1, 1, 1) baseline all the way to grade 3 moves worst-case assurance by nothing at all, because at $E$=1 two of the four fault classes leave no record to examine, so there is nothing for a better-motivated or better-decorrelated auditor to find; the same moves buy 5.8 and 19.2 points of *mean* detection, which is precisely the kind of improvement an average-based score would reward and an adversary would ignore. Only the evidence axis moves the worst case, by 17.4 points. This is the evidence gate acting as a hard threshold, which is a fact about the gate rather than about the aggregation rule: any rule that respects the gate reproduces it, and the grade itself does not register it, since (3, 1, 1), (1, 3, 1) and (1, 1, 3) are all grade 1 while the third is worth 17.4 points of worst-case assurance and the other two none. What the result does show is that a mean-based score would reward the two moves an adversary ignores, and that is the behaviour Proposition 4.4 asks an aggregate to avoid.

*Result 6: the ordering survives the parameters; the magnitudes do not.* The beta factors are declared, so we swept them. Varying the marginal miss rate from 0.02 to 0.30—a fifteen-fold range—and compressing the four substrate beta factors towards their common mean by up to 75%, Kendall's $\tau$ between grade and worst-case assurance across all 64 triples stays within 0.606 to 0.647, and the dominance condition of Proposition 4.2 holds in all twenty settings tested. The strict ordering of the eight named configurations is preserved in all five settings with the declared beta spread and at 25% compression for miss rates up to 0.10; it breaks at 25% compression for miss rates of 0.2 and 0.3, and at 50% or more throughout—the latter is the degenerate case in which the substrate axis has been defined away. Letting the beta factor differ between fault classes by up to a factor of three either way, over 2,000 draws, leaves the ranking of the aggregation rules unchanged: the minimum overstates adversarial assurance in 31.1% of cases and understates in 18.9%, the product in 28.5% and 21.5%, the mean and the weighted sum in 92.0% and 90.4%. Three further checks address the declared values directly. The headline: B's 5.9% ranges from 2.0% to 15.7% as the same-family common-mode share runs over 0.80–0.95 and the internal team's reporting probability over 0.4–0.8, a range in which a same-family auditor remains strongly correlated; the figure is a point in that range, not a prediction. The self-review rule: crediting $P$=0 with a reporting probability of 0.2 instead of zero raises the grade-0 triples from a mean worst case of 1.3% (maximum 9.9%) to 3.2% (maximum 17.2%), reduces the cost of dropping $P$ from the ceiling from 85.9 to 68.7 points, and reverses the near-tie of Result 4: the mean then ranks worst-case assurance above the minimum (0.68 against 0.58) and the grade misorders 134 pairs instead of 62. The comparison of aggregation rules therefore depends on the rule at $P$=0, which is one reason the rule is stated explicitly in the set-up. The top substrate grade: on the verifier-only reading, in which F1 and F4 keep the grade-2 beta factor at $S$=3 (F3 is settled by tamper-evidence either way), C falls from 77.3% to 45.8% worst-case, level with H, D to 50.9%, F to 30.6%, and the catch ratio between $S$=3 and $S$=0 from 8.7 to 5.2. The gap between B and C persists under every reading (0% against 45.8% at least), C's margin over the high-risk minimum does not. What does not survive is the magnitudes. The 5.9% and 77.3% values in Table 5 depend on the chosen parameters. The performance gap persists across the alternative settings examined. B and G have zero worst-case detection under the assumed evidence gates, whereas E has a non-zero, parameter-dependent rate of 8.9% in the baseline setting.

*Result 7: the grade explains most of the assurance, and hides the rest.* Across all 64 triples the grade accounts for 77% of the variance in worst-case assurance ($\eta^2$, Table 6(b)). The residual is the honest cost of an ordinal scale: grade 1 alone spans 0.0% to 51.6%, because configurations F and G are both grade 1 and differ by everything that matters. A reader who needs to separate them must read the triple, not the grade—which is why the protocol requires the triple to be published and treats the grade as a summary of it rather than a replacement for it. The pairwise picture is this: of the 1,158 pairs of triples with different grades, the higher grade has higher worst-case assurance in 70.5%, equal assurance in 24.2% (279 of the 280 ties are pairs that both score zero) and lower assurance in 5.4% (54 of the 62 inversions place a grade-1 triple below a grade-0 one). The grade is informative; it is not reliable enough to carry a threshold on its own, and the tier minima of Table 2 show why. Two triples that fail the high-risk vector minimum outperform it, (1, 3, 2) and (1, 3, 3) at 51.6% against 45.8%, the second being configuration F, which Result 2 holds up as the better buy; the safety-critical minimum (3, 2, 3) scores 50.9%, below F and below (2, 3, 2) at 77.3%. For the low- and high-risk tiers the vector minimum is the same test as $g \geq 1$ and $g \geq 2$, so reading the triple against the vector changes nothing. The cause is not the minimum in particular but fixed thresholds on axes whose levels have not been calibrated against one another: the model says a deficit on one axis can be repaid by a surplus on another, and a threshold cannot express that. The top evidence grade has, in addition, no modelled value in this fault set: no class requires $E$=3, every cell is identical at $E$=2 and $E$=3, and (3, 3, 2) and (3, 3, 3) both score 85.9% at grades 2 and 3. The fault that $E$=3 exists to close is fabrication of records before commitment (Table 2(b), fourth row). Adding it as a fifth class with $e = 3$ sends every triple with $E \leq 2$ to zero worst-case assurance, H included, and leaves twelve of the 64 triples non-zero: whether the high-risk minimum delivers anything in the worst case depends on whether that threat is in scope, and where it is, the safety-critical tier's $E$=3 is what it requires. We therefore read the tier minima as placeholders that encode the model's qualitative priorities—evidence first, substrate second, a change of principal last—and not as calibrated thresholds; the fault-injection study of Section 9 is where thresholds would be set, on measured assurance per fault class rather than on labels.

## 8 Policy Hooks

*EU AI Act, as amended.* Article 12 requires automatic logging for high-risk systems but not tamper-evidence; $E \geq 2$ would give the logs probative value [22]. Article 14 (human oversight) is Step 6; Article 26 places monitoring duties on deployers; Article 43 toggles between conformity assessment by internal control and by a notified body—$P$=1 versus $P$=3—while saying nothing about $S$ or $E$. The Digital Omnibus on AI, proposed in November 2025 and adopted in June 2026, defers Annex III high-risk obligations to 2 December 2027 and Annex I to 2 August 2028 [15, 21], widening the window of voluntary assurance in which the triple lets buyers and insurers compare offerings.

*Standards.* ISO/IEC 42006:2025, which specifies who may audit and certify AI management systems [35, 36], governs impartiality—$P$—and is silent on $S$ and $E$; certification bodies could report the triple now.

*UK risk management.* The AI Risk Management Toolkit published in September 2026 implements the Orange Book's risk process for public-sector AI, alongside a roadmap to professionalise third-party AI assurance [17] and a gap analysis flagging agentic-system security as under-studied [19]. It is self-assessment, and most of it needs no audit; but several treatment options do, each specified as a binary (Table 7). A team may therefore record them as applied, and lower its residual-risk score, while the independent testers run the auditee's model family and the preserved logs stay rewritable by the operator—grade 0 here. Teams are also asked to estimate risk likelihood partly by model analysis: where the estimating model shares a substrate with the system scored, Eq. (2) says the estimate carries less independent information than the count of checks implies. Recording ($P$, $S$, $E$) against each such treatment in the risk workbook costs three integers and makes the residual-risk score mean what it says.

Table 7. Elements of the UK AI Risk Management Toolkit [18] whose efficacy depends on an unstated degree of independence, and what the triple supplies.

| Toolkit element | Axis | As specified | What the grade adds |
|---|---|---|---|
| Adversarial testing of accuracy and performance by external experts | $P$, $S$ | Independent of the AI project team | Whether the testers share the auditee's substrate; $S \geq 2$ |
| Test and evaluation by separate teams, to counter groupthink | $P$, $S$ | Separate individuals or teams | Same-family testers inherit the blind spots they are there to break |
| Auditability: process traceability, data provenance, logging | $E$ | Logging required; tamper-evidence not | $E \geq 2$ gives the log probative value |
| Materials preserved for forensic, regulatory and legal review | $E$ | Preservation required | Preserved but alterable is not evidence; $E \geq 2$ |
| Likelihood estimated by model analysis | $S$ | Method listed; correlation not addressed | $n_{\text{eff}}$ discount for a same-substrate estimator, Eq. (2) |
| Disclosure of audits; accountability metrics | $P$,$S$,$E$ | Audits disclosed | The triple is the metadata that makes a disclosed audit interpretable |

*Audit regulators.* The Financial Reporting Council's review of the six largest UK firms found no formal monitoring of the audit-quality impact of their automated tools and, at all but one firm, no indicators for them [24]; the PCAOB stresses continued human supervision of generative-AI output [53]. Both could require the triple whenever an agent contributes evidence, treating a same-substrate agent as $S$=0 for reliance.

*Statutory precedent.* New York City's Local Law 144 has mandated independent bias audits of automated employment decision tools since 2023 [50]; its definition of an independent auditor speaks only to $P$, and adding minimum $S$ and $E$ would be a modest drafting change.

## 9 Limitations and Validation Plan

Attestation proves which code and configuration ran, not that they behaved correctly: $E$=3 is necessary for trustworthy evidence, not sufficient for a correct opinion, and inherits the enclave's trust base and side-channel history [13]; verifiable inference at frontier scale is not production-ready [60]. Substrate lineage depends on disclosure; vendors will contest the $S \leq 1$ cap for undisclosed lineage, but it is the conservative default. Equations (1)–(2) assume exchangeable auditors and a single $\rho$; real fault classes differ, which is why Table 4 reasons per class. The common-shock model buys the monotonicity of Proposition 4.1 at the price of two assumptions a critic should press on: that shared components induce common-mode failure independently of one another, which overstates $\gamma_D$ where two shared components fail through the same mechanism, and that the per-component rates $\gamma_c$ are estimable at all—today they are not, so Proposition 4.2 yields a qualitative dominance condition rather than a procurement threshold. The calibration transfers a figure measured on evaluation panels to auditing, which is a hypothesis the study below is designed to test, not a result; and, as Result 3 shows, even a correct transfer would fix only an upper bound on the floor of Proposition 4.2, because correlated misses need no shared component and pairwise data cannot tell a common-shock floor from a difficulty-function tail. The grades are ordinal and the min rule discards information. Section 7 sharpens what we can claim for it: the minimum is not uniformly the better *predictor* of assurance—against a worst-case target it is level with a mean and marginally below a product—and its case rests on the argument of Proposition 4.4, its readability, and understating less than the other rules that rarely overstate; rarely overstating on its own does not single it out (Result 4), and the minimum of three separate ordinal scales is not invariant to relabelling one of them. A reader who wants an expected-yield estimate should not use the grade for it. The numerical study validates the model, not the world. Its parameters are declared rather than measured, and Result 6 is explicit about what that costs: the rank relationship between grade and assurance is stable across a fifteen-fold sweep of the miss rate and a 75% compression of the substrate beta factors, but the magnitudes are artefacts of the chosen values and should not be quoted as expected detection rates for any real system; the headline figure moves between 2.0% and 15.7% over a plausible range of its two declared inputs. Result 7 adds a second caveat that the grade itself carries: it fixes 77% of the variance in worst-case assurance and leaves a within-grade spread of up to 51.6 points, so the triple is the reportable object and the grade only a summary of it; and the tier minima are placeholders, since the model misorders named configurations across them and contains no fault that the top evidence grade is needed for. Three things the model does not contain should be named. It has no term for false findings: an auditor that raises exceptions a human must resolve has a cost, and a cascade analysis of two checks shows that a second check whose misses duplicate the first and which has any friction at all makes the error rate among items let through strictly worse [63]; a rubric that scores only missed faults will always favour more auditing, so a false-finding rate per axis is the next term to add to the model; the protocol already asks for the measured rate at Step 7. It does not model the human auditor's own detection, only the team's reporting probability (Section 7, Set-up), so it says nothing about an audit function that re-performs checks itself. And the protocol assumes that lineage, where disclosed, is disclosed truthfully and completely; the case of a guardrail package that shipped with most of its rules silently unloaded [62] shows that lineage must record the configuration in force at run time, which attestation of the package would not by itself have caught. The empirical study this calls for is well defined: inject F1–F4 and the pre-commitment fabrication class of Result 7 into benchmark agents [1, 16, 38] audited by agents at $S = 0, \dots, 3$, estimate the beta factor per fault class instead of assuming it—reported with intervals for each auditor–auditee pair, since the dependence the substrate axis stands for can be estimated only where the presence of a fault is known independently of the system under audit [63]—test whether the lineage grade predicts those estimates, include same-substrate panels of increasing size, because three reviewers is the first size at which a common-shock floor and a difficulty tail separate (Result 3) and a three-reviewer rate alone will not separate a small floor from none, and compare measured detection against the model's predictions; a cheaper companion study scores the triple retrospectively against treatments already recorded in public-sector risk workbooks [18]. Until that is done the numbers in Section 7 should be read as consequences of the model and nothing more. Grade-3 principals barely exist; agent-governance law [43] must catch up.

One external application exists. A provider of verification infrastructure graded its own published estate at $P$=2 and $E$=2 [31]; its substrate position follows from the data-path rule of Section 4 once the independence of its reads from the subjects' delivery path is stated. A self-applied grade is a claim at $P$=0 about $P$=2, as its author notes, and we report it as a worked instance rather than a determination. The same provider has published a proficiency study on its instrument—2,940 evaluations across 60 cases and four fault families, a detection floor per family, and a negative control of 60 clean tracks returning no false flag, with a sample of the checks re-runnable by a stranger in one command [31]. That is the shape of measurement Step 7 asks for, and the negative control is the part a model of sensitivity alone cannot supply. It is not a measurement of agentic audit: it injects no prompt injection, its four fault families are not F1–F4, and nothing in it is a value for the beta factor, which remains unmeasured.

## 10 Conclusion

Independence was never meant to be a checkbox, and for agentic systems it cannot be. Two agents on the same substrate confirming each other is one opinion, not two—Eq. (2) makes the arithmetic explicit; an audit built on self-reported logs is a narrative, not evidence. On the model of Section 4, an internal audit that does everything an audit function is normally asked to do, but on the auditee's model family and the provider's logs, surfaces nothing at all in the worst case—a conclusion that follows from the evidence gate and not from any declared parameter. Grading independence along principal, substrate and evidence, and publishing the triple with every opinion, costs little, needs no new law and reveals how much an "independent" audit of an agent actually shows.

## Declarations

### Ethics.

The study involves no human or animal subjects, no personal data and no user research; the numerical study computes closed forms of a declared model, and its only random draws are implementation checks. The protocol itself touches personal data only indirectly: the tool-call records that carry evidence grades may contain personal data, so a deployment applying Step 4 should minimise and retain them under the applicable data-protection regime rather than logging indiscriminately in pursuit of a higher $E$. There is a real tension here—higher evidence grades mean more retained, more durable records—and a deployment should resolve it by scoping tamper-evidence to load-bearing records rather than to everything the agent does.

### Acknowledgements.

The author thanks Taiko Toeda (MOBIUS LLC, Tokyo), who replicated the numerical study from the public package, and Joel Hillier (Certisyn, Inc.), who applied the protocol to a deployed verification estate; their results are cited where they are used.

### Conflicts of interest.

The author holds academic appointments at Keele University and the University of Liverpool. No funder had any role in the design or conclusions of this work. The self-review risk the paper analyses applies to the paper: a framework proposed by someone with an interest in the assurance market should be read with the incentive in view, which is one reason the rubric is specified so that a third party can apply it without the author's involvement.

### Data availability.

The numerical study of Section 7 is a small Python package (independence-grading, version 1.1.0) with three experiment drivers—one for Results 1 to 4, one for the ablation, parameter sweep, full grid, heterogeneity and the alternative-reading checks of Results 5 to 7, and one for the identifiability analysis of Result 3—together with twenty-nine unit tests that check the closed forms, the gates, the mechanical settlement of F3, the zero at $P$=0, the aggregation rules and the identifiability result, and that verify the implementation against the Monte Carlo draws. It depends on NumPy and SciPy only and reproduces every number reported here in under two minutes; every number is a closed form, and the fixed seed governs only the implementation checks and the heterogeneous-beta draws. It is provided with this submission and deposited on Zenodo (10.5281/zenodo.22768569). There is no data set: the study computes the model's closed forms and depends on no external input.